# Efficient Natural Language Response Suggestion for Smart Reply

MATTHEW HENDERSON, RAMI AL-RFOU, BRIAN STROPE, YUN-HSUAN SUNG,
LÁSZLÓ LUKÁCS, RUIQI GUO, SANJIV KUMAR, BALINT MIKLOS, and
RAY KURZWEIL, Google

This paper presents a computationally efficient machine-learned method for natural language response suggestion. Feed-forward neural networks using n-gram embedding features encode messages into vectors which are optimized to give message-response pairs a high dot-product value. An optimized search finds response suggestions. The method is evaluated in a large-scale commercial e-mail application, *Inbox by Gmail*. Compared to a sequence-to-sequence approach, the new system achieves the same quality at a small fraction of the computational requirements and latency.

Additional Key Words and Phrases: Natural Language Understanding; Deep Learning; Semantics; Email

## 1 INTRODUCTION

Applications of natural language understanding (NLU) are becoming increasingly interesting with scalable machine learning, web-scale training datasets, and applications that enable fast and nuanced quality evaluations with large numbers of user interactions.

Early NLU systems parsed natural language with hand-crafted rules to explicit semantic representations, and used manually written state machines to generate specific responses from the output of parsing [18]. Such systems are generally limited to the situations imagined by the designer, and much of the development work involves writing more rules to improve the robustness of semantic parsing and the coverage of the state machines. These systems are brittle, and progress is slow [31]. Eventually adding more parsing rules and response strategies becomes too complicated for a single designer to manage, and dependencies between the rules become challenging to coordinate across larger teams. Often the best solution is to keep the domains decidedly narrow.

Statistical systems can offer a more forgiving path by learning implicit trade-offs, generalizations, and robust behaviors from data. For example, neural network models have been used to learn more robust parsers [14, 24, 29]. In recent work, the components of task-oriented dialog systems have been implemented as neural networks, enabling joint learning of robust models [7, 26, 27]. However these methods all rely on either an explicit semantic representation or an explicit representation of the task, always hand-crafted.

End-to-end systems avoid using hand-crafted explicit representations, by learning to map to and from natural language via implicit internal vector representations [19, 25]. Such systems avoid the unnecessary contraints and bottlenecks inevitably imposed by the system designer. In that context, natural language understanding might be evaluated less in terms of an explicit semantic representation, and more by the utility of the system itself. The system shows evidence of understanding when it offers useful responses.

Such end-to-end tasks are difficult: systems not only need to learn language but also must learn to do something useful with it. This paper addresses the task of suggesting responses in human-to-human conversations. There are further challenges that arise when building an end-to-end dialog system, i.e. a computer agent that interacts directly with a human user. Dialog systems must learn effective and robust interaction strategies, and goal-oriented systems may need to interact with discrete external databases. Dialog systems must also learn to be consistent throughout the course of a dialog, maintaining some kind of memory from turn to turn.

Machine learning requires huge amounts of data, and lots of helpful users to guide development through live interactions, but we also need to make some architectural decisions, in particular how to represent natural language text.

Neural natural language understanding models typically represent words, and possibly phrases, sentences, and documents as implicit vectors. Vector representations of words, or *word embeddings*, have been widely adopted, particularly since the introduction of efficient computational learning algorithms that can derive meaningful embeddings from unlabeled text [15, 17, 20].

Though a simple representation of a sequence of words can be obtained by summing the individual word embeddings, this discards information about the word ordering. The sequence-to-sequence (*Seq2Seq*) framework uses recurrent neural networks (RNNs), typically long short-term memory (LSTM) networks, to encode sequences of word embeddings into representations that depend on the order, and uses a decoder RNN to generate output sequences word by word. This framework provides a direct path for end-to-end learning [23]. With attention mechanisms and more layers, these systems are revolutionizing the field of machine translation [28]. A similar system was initially used to deployed Google's *Smart Reply* system for *Inbox by Gmail* [11].

While Seq2Seq models provide a generalized solution, it is not obvious that they are maximally efficient, and training these systems can be slow and complicated. Also they are derived as a generative model, and so using them to rank a fixed set of responses (as in the context of Smart Reply) requires extra normalization to bias the system away from common responses.

In a broader context, Kurzweil's work outlines a path to create a simulation of the human neocortex (the outer layer of the brain where we do much of our thinking) by building a hierarchy of similarly structured components that encode increasingly abstract ideas as sequences [12]. Kurzweil provides evidence that the neocortex is a self-organizing hierarchy of modules, each of which can learn, remember, recognize and/or generate a sequence, in which each sequence consists of a sequential pattern from lower-level modules. Longer relationships (between elements that are far away in time or spatial distance) are modeled by the hierarchy itself. In this work we adopt such a hierarchical structure, representing each sequential model as a feed-forward vector computation (with underlying sequences implicitly represented using n-grams). Whereas a long short-term memory (LSTM) network could also model such sequences, we don't need an LSTM's ability to directly encode long-term relationships (since the hierarchy does that) and LSTMs are much slower than feed-forward networks for training and inference since the computation scales with the length of the sequence.

Similarly, the work on *paragraph vectors* shows that word embeddings can be back-propagated to arbitrary levels in a contextual hierarchy [13]. Machines can optimize sentence vectors, paragraph vectors, chapter vectors, book vectors, author vectors, and so on, with simple back-propagation and computationally efficient feed-forward networks.

Putting a few of these ideas together, we wondered if we could predict a sentence using only the sum of its n-gram embeddings. Without the ordering of the words, can we use the limited sequence information from the n-grams, and the redundancy of language, to recreate the original word sequence? With a simple RNN as a decoder, our preliminary experiments showed perplexities of around 1.2 over a vocabulary of hundreds of thousands of words. A lot of the sequence information remains in this simple n-gram sentence representation. As a corollary, a hierarchy built on top of n-gram representations could indeed adequately represent the increasingly abstract sequences underlying natural language. Networks built on n-gram embeddings such as those presented in this paper (see section 4) are computationally inexpensive relative to RNN and convolutional network [6, 30] encoders.

To make sure there is enough data and the necessary live feedback from users, we train on the anonymized Gmail data that was used in Kannan et al. [11], and use our models to give *Smart Reply* response suggestions to users of *Inbox by Gmail* (see figure 1). Smart Reply provides a real world application in which we can measure the quality of our response suggestion models.

Just as in Kannan et al. [11], we consider natural language response suggestion from a fixed set of candidates. For efficiency, we frame this as a search problem. Inputs are combined with potential responses using final dot products to enable precomputation of the "response side" of the system. Adding deep layers and delaying combination between input and responses encourages the network to derive implicit semantic representations of the input and responses– if we assume that the best way to predict their relationships is to understand them.

We precompute a minimal hierarchy of deep feed-forward networks for all potential responses, and at runtime propagate only the input through the hierarchical network. We use an efficient nearest-neighbor search of the hierarchical embeddings of the responses to find the best suggestions.

## 2 PROBLEM DEFINITION

The Smart Reply system gives short response suggestions to help users respond quickly to emails. Emails are processed by the system according to the pipeline detailed in figure 2.

The decision of whether to give suggestions is made by a deep neural network classifier, called the *triggering* model. This model takes various features of the received email, including a word n-gram representation, and is trained to estimate the probability that the user would type a short reply to the input email, see Kannan et al. [11]. If the output of the triggering model is above a threshold, then Smart Reply will give $m$ (typically 3) short response suggestions for the email. Otherwise no suggestions are given. As a result, suggestions are not shown for emails where a response is not likely (e.g. spam, news letters, and promotional emails), reducing clutter in the user interface and saving unnecessary computation.

The system is restricted to a fixed set of response suggestions, $R$, selected from millions of common messages. The *response selection* step involves searching for the top $N$ (typically around 100) scoring responses in $R$ according to a response selection model $P(y \mid x)$. The output of response selection is a list of suggestions $(y_1, y_2, \ldots, y_N)$ with $y_i \in R$ ordered by their probability. Kannan et al. [11] used a sequence-to-sequence model for $P(y \mid x)$ and used a beam search over the prefices in $R$ (see section 3). This paper presents a feed-forward neural network model for $P(y \mid x)$, including a factorized *dot-product* model where selection can be performed using a highly efficient and accurate approximate search over a precomputed set of vectors, see section 4.

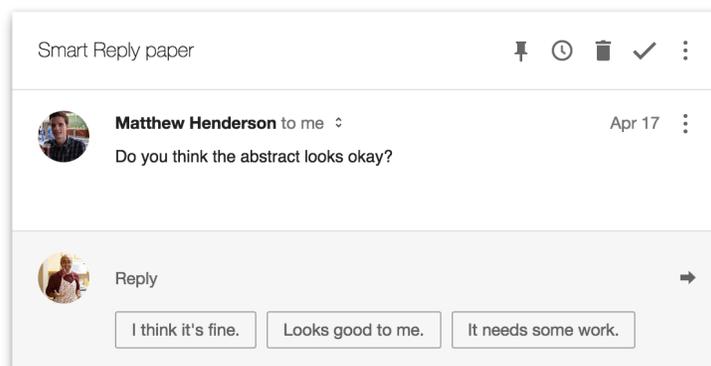

Fig. 1. Our natural language understanding models are trained on email data, and evaluated in the context of the *Smart Reply* feature of *Inbox by Gmail*, pictured here.

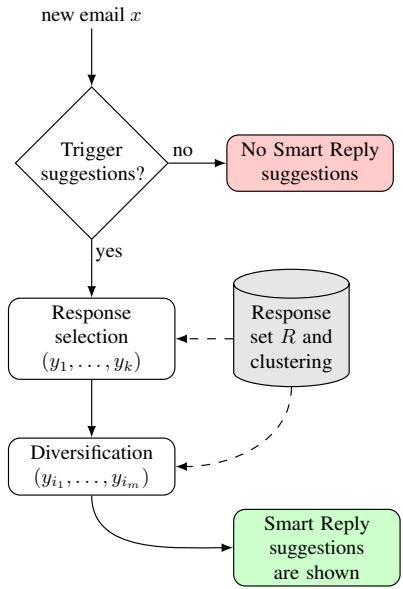

Fig. 2. The Smart Reply pipeline. A received email is run through the triggering model that decides whether suggestions should be given. Response selection searches the response set for good suggestions. Finally, diversification ensures diversity in the final set shown to the user. This paper focuses on the response selection step.

Finally the *diversification* stage ensures diversity in the final $m$ response suggestions. A clustering algorithm is used to omit redundant suggestions, and a labeling of $R$ is used to ensure a negative suggestion is given if the other two are affirmative and vice-versa. Full details are given in Kannan et al. [11].

## 3  BASELINE SEQUENCE-TO-SEQUENCE SCORING

The response selection model presented in Kannan et al. [11] is a long short-term memory (LSTM) recurrent neural network [8] – an application of the sequence-to-sequence learning framework (*Seq2Seq*) [23].

The input email $x$ is tokenized into a word sequence $(x_1, \ldots, x_m)$ and the LSTM computes the conditional probability over a response sequence $y = (y_1, \ldots, y_n)$ as:

$$
\begin{aligned}
P(y \mid x) &= P(y_1, \ldots, y_n \mid x_1, \ldots, x_m) \\
&= \prod_{i=1}^{n} P_{\text{LSTM}}(y_i \mid x_1, \ldots, x_m, y_1, \ldots, y_{i-1})
\end{aligned}
$$

where $P_{\text{LSTM}}$ is the output of the word-level LSTM. The LSTM is trained to maximize the log-probability according to $P(y \mid x)$ of the training data (a large collection of emails and responses, see section 5.1). At inference time, likely responses from the candidate set $R$ are found using a beam search that is restricted to the prefix trie of $R$. The time complexity of this search is $O(|x| + b|y|)$ where $b$ is the beam width and should be scaled appropriately with $|R|$. This search dominates the computation of the original Smart Reply system.

## 4  FEEDFORWARD APPROACH

Rather than learning a generative model, we investigate learning a feedforward network to score potential responses.

Recall the goal of response selection is to model $P(y \mid x)$, which is used to rank possible responses $y$ given an input email $x$. This probability distribution can be written as:

$$P(y \mid x) = \frac{P(x, y)}{\sum_k P(x, y_k)} \tag{1}$$

The joint probability of $P(x, y)$ is estimated using a learned neural network scoring function, $S$ such that:

$$P(x, y) \propto e^{S(x,y)} \tag{2}$$

Note that the calculation of equation 1 requires summing over the neural network outputs for all possible responses $y_k$. (This is only an issue for training, and not inference since the denominator is a constant for any given $x$ and so does not affect the $\arg\max$ over $y$). This is prohibitively expensive to calculate, so we will approximate $P(x)$ by sampling $K$ responses including $y$ uniformly from our corpus during training:

$$P_{\text{approx}}(y \mid x) = \frac{P(x, y)}{\sum_{k=1}^{K} P(x, y_k)} \tag{3}$$

Combining equations 2 and 3 gives the approximate probability of the training data used to train the neural networks:

$$P_{\text{approx}}(y \mid x) = \frac{e^{S(x,y)}}{\sum_{k=1}^{K} e^{S(x,y_k)}} \tag{4}$$

The following subsections show several scoring models; how to extend the models to multiple features; how to overcome bias introduced by the sampling procedure; and an efficient search algorithm for response selection.

### 4.1 N-gram Representation

To represent input emails $x$ and responses $y$ as fixed-dimensional input features, we extract n-gram features from each. During training, we learn a $d$-dimensional embedding for each n-gram jointly with the other neural network parameters. To represent sequences of words, we combine n-gram embeddings by summing their values. We will denote this bag of n-grams representation as $\Psi(x) \in \mathbb{R}^d$. This representation is quick to compute and captures basic semantic and word ordering information.

### 4.2 Joint Scoring Model

Figure 3a shows the joint scoring neural network model that takes the bag of n-gram representations of the input email $x$ and the response $y$, and produces a scalar score $S(x, y)$. This deep neural network can model complex joint interactions between input and responses in its computation of the score.

### 4.3 Dot-Product Scoring Model

Figure 3b shows the structure of the dot-product scoring model, where $S(x, y)$ is factorized as a dot-product between a vector $\mathbf{h}_x$ that depends only on $x$ and a vector $\mathbf{h}_y$ that depends only on $y$. This is similar to Deep Structured Semantic Models, which use feedforward networks to project queries and documents into a common space where the relevance of a document given a query is computed as the cosine distance between them [9].

While the interaction between features is not as direct as the joint scoring model (see section 4.2), this factorization allows us to calculate the representation of the input $x$ and possible responses $y$

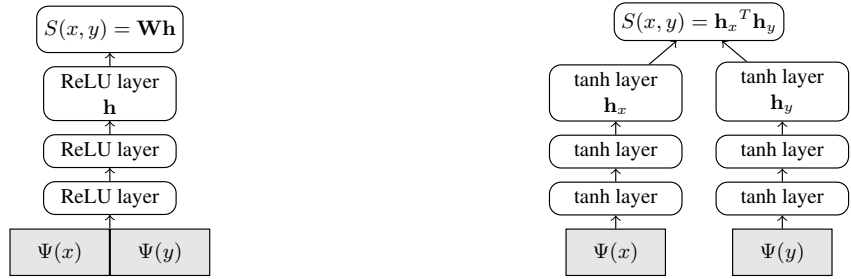

(a) A neural network that calculates a score between emails and their responses. Rectified Linear Unit (ReLU) layers are used to reduce the $(2d)$-dimensional concatenation of the bag of n-gram representations to a scalar $S(x, y)$.

(b) Dot-product architecture, where a tower of tanh activation hidden layers encodes $x$ to $\mathbf{h}_x$ and a separate tower encodes $y$ to $\mathbf{h}_y$ such that the score $S(x, y)$ is the dot-product $\mathbf{h}_x^T \mathbf{h}_y$.

Fig. 3. Feedforward scoring models that take the n-gram representation of an email body and a response, and compute a score.

independently. In particular, the representations of the response set $R$ can be precomputed. Then searching for response suggestions reduces to encoding a new email $x$ in a simple feed-forward step to the vector $\mathbf{h}_x$, and then searching for high dot-product scoring responses in the precomputed set (see section 4.7).

It is also efficient to compute the scores $S(x_i, y_j)$ for all pairs of inputs and responses in a training batch of $n$ examples, as that requires only an additional matrix multiplication after computing the $\mathbf{h}_{x_i}$ and $\mathbf{h}_{y_i}$ vectors. This leads to vastly more efficient training with multiple negatives (see section 4.4) than is possible with the joint scoring model.

### 4.4 Multiple Negatives

Recall from section 4 that a set of $K$ possible responses is used to approximate $P(y \mid x)$ – one correct response and $K - 1$ random *negatives*. For efficiency and simplicity we use the responses of other examples in a training batch of stochastic gradient descent as negative responses. For a batch of size $K$, there will be $K$ input emails $\mathbf{x} = (x_1, \ldots, x_K)$ and their corresponding responses $\mathbf{y} = (y_1, \ldots, y_K)$. Every reply $y_j$ is effectively treated as a negative candidate for $x_i$ if $i \neq j$. The $K - 1$ negative examples for each $x$ are different at each pass through the data due to shuffling in stochastic gradient descent.

The goal of training is to minimize the approximated mean negative log probability of the data. For a single batch this is:

$$\begin{aligned}
\mathcal{J}(\mathbf{x}, \mathbf{y}, \theta) \\
&= -\frac{1}{K} \sum_{i=1}^{K} \log P_{\text{approx}}(y_i \mid x_i) \\
&= -\frac{1}{K} \sum_{i=1}^{K} \left[ S(x_i, y_i) - \log \sum_{j=1}^{K} e^{S(x_i, y_j)} \right]
\end{aligned} \quad (5)$$

using equation 4, where $\theta$ represents the word embeddings and neural network parameters used to calculate $S$. Note that this loss function is invariant to adding any function $f(x)$ to $S(x, y)$, so

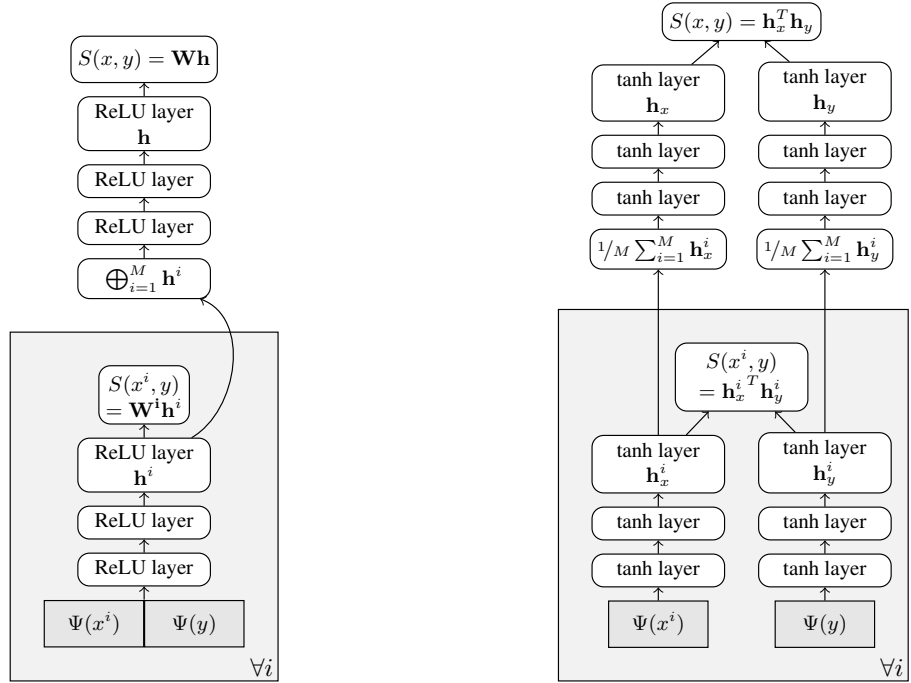

(a) Joint scoring model using multiple features of the input email $x^i$. A subnetwork scores the response using each feature alone, before the top-level hidden representations $\mathbf{h}^i$ are concatenated ($\bigoplus_{i=1}^{M} \mathbf{h}^i$) and then used to compute the final score. This is an application of the multi-loss architecture from Al-Rfou et al. [2].

(b) Dot-product scoring model with multiple input features $x^i$. This is a novel setup of the multi-loss architecture, whereby the feature-level scores $S(x^i, y)$ and the final score $S(x, y)$ are computed as a dot-product between the parallel input and response sides.

Fig. 4. Scoring models that use multiple features of the input email.

$S(x, y)$ is learned up to an additive term that does not affect the $\arg\max$ over $y$ performed in the inference time search.

### 4.5 Incorporating Multiple Features

There is structure in emails that can be used to improve the accuracy of scoring models. We follow the *multi-loss* architecture of Al-Rfou et al. [2] to incorporate additional features beyond the message body, for example the subject line. Figure 4 shows the multi-loss architecture applied to both the joint and dot-product scoring models.

The multi-loss networks have a sub-network for each feature of the email, which are trained to independently score candidate responses using that feature alone. The highest level hidden layer of the sub-network is used in a final sub-network that is trained to combine the information from all the features and give a final score. This hierarchical structure results in models that learn how to use each feature faster than a network that sees all the features at once, and also allows for learning deeper networks than is otherwise possible [2].

Formally, denote the $M$ features of an input email $x$ as $x^1$, ..., $x^M$. Then for each $i$, a sub-network produces a hidden vector representation $\mathbf{h}^i$, and a score of the response $y$ using only $x^i$, $S(x^i, y)$. Denoting $(x_1^i, \ldots, x_K^i)$ as $\mathbf{x}^i$, a loss function $\mathcal{J}(\mathbf{x}^i, \mathbf{y}, \theta)$ encourages $S(x^i, y)$ to be high for corresponding pairs in the training batch, and low for the random pairs. The second stage of the network produces a final score $S(x, y)$ that is a function of all of the $\mathbf{h}^i$ vectors. The network is trained end-to-end with a single loss:

$$\mathcal{J}(\mathbf{x}, \mathbf{y}, \theta) + \sum_{i=1}^{M} \mathcal{J}(\mathbf{x}^i, \mathbf{y}, \theta)$$

| **Message**: Did you manage to print the document? ||
| --- | --- |
| **With response bias** | **Without response bias** |
| – Yes, I did. | – It's printed. |
| – Yes, it's done. | – I have printed it. |
| – No, I didn't. | – Yes, all done. |

Table 1. Examples of Smart Reply suggestions with and without the response bias. Without biasing, the model prefers responses that are very closely related to the input email, but are less likely to be chosen than the more generic yes/no responses.

Note that the final score produced by the multi-loss dot-product model (figure 4b) is a dot-product of a vector $\mathbf{h}_x$ that depends only on the input $x$, and a vector $\mathbf{h}_y$ that depends only on the response $y$, as in the single-feature case. As a result, it can still be used for the fast vector search algorithm described in section 4.7, and training with multiple negatives remains efficient.

For the multi-loss joint scoring model, the input feature vector for the final sub-network is the concatenation of the $\mathbf{h}^i$ vectors and therefore scales with the number of features, leading to a computational bottleneck. For the dot-product scoring model, the hidden layer representations are learned such that they are meaningful vectors when compared using a dot product. This motivates combining the representations for the final sub-network using vector arithmetic. The features extracted from the input email, $x^i$, are averaged ($1/M \sum_{i=1}^{M} \mathbf{h}_x^i$), as are the response representations learned from different sub-networks ($1/M \sum_{i=1}^{M} \mathbf{h}_y^i$), before being passed to the final neural network layers. While this choice may constrain the representations learned by each sub-network, and may limit the ability of the final sub-network to differentiate information from different features, it also encourages them to exist in the same semantic space.

### 4.6 Response Biasing

The discriminative objective function introduced in section 4.4 leads to a biased estimation of the denominator in equation (1). Since our negative examples are sampled from the training data distribution, common responses with high prior likelihood appear more often as negative examples. In practice, we observed that this bias leads to models that favor specific and long responses instead of short and generic ones. To encourage more generic responses, we bias the responses in $R$ using a score derived from the log likelihood of the response as estimated using a language model. Our final score $S_f(x, y)$ of any input email response pair is calculated as:

$$S_f(x, y) = S_m(x, y) + \alpha \log P_{\text{LM}}(y) \qquad (6)$$

where $S_m$ is the score calculated by our trained scoring model, $P_{\text{LM}}(y)$ is the probability of $y$ according to the language model, and $\alpha$ is tuned with online experiments. Note that the additional term is dependent only on $y$, and so can be precomputed for every response in $R$ prior to inference time.

Table 1 demonstrates the effect of including the response bias using an example email.

### 4.7 Hierarchical Quantization for Efficient Search

At inference time, given an input email $x$, we use the dot-product scoring model to find response suggestions $y \in R$ with the highest scores $S(x, y)$, where the scoring function is the dot-product: $S(x, y) = \mathbf{h}_x^T \mathbf{h}_y$[1]. The problem of finding datapoints with the largest dot-product values is sometimes called Maximum Inner Product Search (MIPS). This is a research topic of its own and is also useful for inference in neural networks with a large number of output classes.

Maximum Inner Product Search is related to nearest neighbor search (NNS) in Euclidean space, but comes with its own challenges because the dot-product "distance" is non-metric and many classical approaches such as KD-trees cannot be applied directly. For more background, we refer readers to the relevant works of [3, 5, 21, 22]. In the Smart Reply system, we need to keep very high retrieval recall (for example $> 99\%$ in top-30 retrieval). However, many of the existing methods are not designed to work well in the high recall regime without slowing down the search considerably. To achieve such high recall, hashing methods often require a large number of hash bits and tree methods often need to search a large number of leaves.

In this work, we use a hierarchical quantization approach to solve the search problem. For our use case, the responses $y$ come from a fixed set $R$ and thus the $\mathbf{h}_y$ vectors are computed ahead of inference time. Unlike the previous work in [5], we propose a hierarchical combination of vector quantization, orthogonal transformation and product quantization of the transformed vector quantization residuals. Our hierarchical model is based on the intuition that data in DNN hidden layers often resemble low dimensional signal with high dimensional residuals. Vector quantization is good at capturing low dimensional signals. Product quantization works by decomposing the high-dimensional vectors into low-dimensional subspaces and then quantizing them separately [4]. We use a learned rotation before product quantization as it has been shown to improve quantization error [16].

Specifically, $\mathbf{h}_y$ is approximated by a hierarchical quantization $HQ(\mathbf{h}_y)$, which is the sum of the vector quantization component $VQ(\mathbf{h}_y)$ and the residuals. A learned orthogonal transformation $\mathbf{R}$ is applied to the residual, followed by product quantization.

$$\mathbf{h}_y \approx HQ(\mathbf{h}_y) = VQ(\mathbf{h}_y) + \mathbf{R}^T PQ(r_y),$$
$$\text{where} \quad r_y = \mathbf{R}(\mathbf{h}_y - VQ(\mathbf{h}_y))$$

Here, given a vector quantization codebook $\mathbf{C_{VQ}}$, product quantization codebooks of $\{\mathbf{C_{PQ}}^{(k)}\}$ for each of the subspaces $k$, and the learned orthogonal matrix $\mathbf{R} \in \mathbb{R}^{d \times d}$, the vector quantization of $\mathbf{h}_y$ is $VQ(\mathbf{h}_y) = \arg\min_{c \in \mathbf{C_{VQ}}} ||\mathbf{h}_y - c||^2$. The product quantization of the rotated residual $r_y$ is computed by first dividing the rotated residuals $r_y$ into $\mathcal{K}$ subvectors $r_y^{(k)}$, $k = 1, 2, \cdots, \mathcal{K}$, and then quantizing the subvectors independently by vector quantizers $\mathbf{C_{PQ}}^{(k)}$:

$$PQ^{(k)}(r_y^{(k)}) = \arg\min_{s \in \{\mathbf{C_{PQ}}^{(k)}\}} ||s - r_y^{(k)}||^2.$$

Finally the full product quantization $PQ(r_y)$ is given by the concatenation of the quantization in each subspace:

$$PQ(r_y) = \begin{pmatrix} PQ^{(1)}(r_y^{(1)}) \\ PQ^{(2)}(r_y^{(2)}) \\ \vdots \\ PQ^{(\mathcal{K})}(r_y^{(\mathcal{K})}) \end{pmatrix}, \quad r_y = \begin{pmatrix} r_y^{(1)} \\ r_y^{(2)} \\ \vdots \\ r_y^{(\mathcal{K})} \end{pmatrix}$$

---

[1] The bias term, $\alpha \log P_{\text{LM}}(y)$, can be included in the dot product e.g. by extending the $\mathbf{h}_x$ vector with $\{\alpha\}$ and the $\mathbf{h}_y$ vector with $\{\log P_{\text{LM}}(y)\}$

At training time, the codebook for vector quantization, $\mathbf{C_{VQ}}$, codebooks for product quantization $\mathbf{C_{PQ}^{(\cdot)}}$, and the rotation matrix $\mathbf{R}$ are jointly learned by minimizing the reconstruction error of $\mathbf{h}_y - HQ(\mathbf{h}_y)$ with stochastic gradient descent (SGD). At inference time, prediction is made by taking the candidates with the highest quantized dot product, i.e.

$$\mathbf{h}_x^T VQ(\mathbf{h}_y) + (\mathbf{R}\mathbf{h}_x)^T PQ(r_y)$$

The distance computation can be performed very efficiently without reconstructing $HQ(\mathbf{h}_y)$, instead utilizing a lookup table for asymmetric distance computation [10]. Furthermore, the lookup operation is carried out in register using SIMD (single instruction, multiple data) instructions in our implementation, providing a further speed improvement.

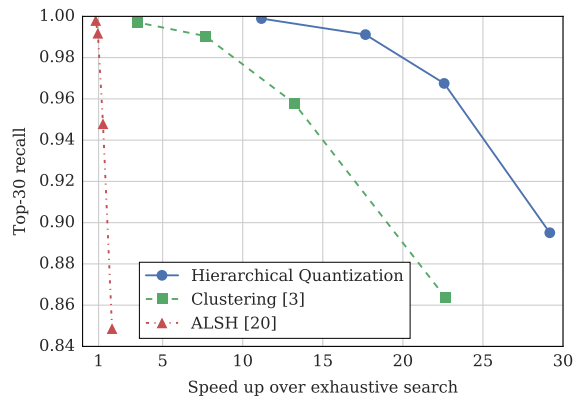

Fig. 5. Evaluation of retrieval speed vs. recall of top-30 neighbors with maximum dot product. The curve is produced by varying the number of approximate neighbors retrieved by our hierarchical quantization method and by Asymmetric LSH [22], and varying the number of leaves searched by the clustering algorithm of [3].

We summarize the speed-recall evaluation of using different approximate MIPS algorithms in figure 5. The y-axis shows the recall of the top-30 retrieved responses, where the ground truth is computed by exhaustive search. The x-axis shows the speed up factor with respect to exhaustive search. Therefore, exhaustive search achieves a recall of 100% with a speed up factor of 1. Our algorithm achieves 99.89% recall with a speed-up factor over 10, outperforming the baselines of [3, 22].

## 5 EVALUATION
### 5.1 Experimental Setup

*Data*. Pairs of emails and their responses are sampled from user data to create datasets for training and testing the feedforward response scoring models. In total around 300M pairs are collected. The data is split uniformly at random into two disjoint sets of 95% and 5%, which constitute the training and test sets respectively.

All email data (raw data, preprocessed data and training/evaluation data) is encrypted. Engineers can only inspect aggregated statistics on anonymized sentences that occurred across many users and do not identify any user.

Language identification is run on the emails, and only English language emails are kept. The subject lines and message bodies are tokenized into word sequences, from which n-gram features are extracted. Infrequent words, URLs, email addresses, phone numbers etc. are replaced with special tokens. Quoted text arising from replying and forwarding is also removed. We used hundreds of thousands of the most frequent n-grams as features to represent the text.

*Training*. Each of our DNN sub-networks consists of 3 hidden layers of sizes 500, 300, 100 in the case of the joint scoring models and 300, 300, 500 for the dot-product models . The embedding dimensionality $d$ of our n-grams is 320. We train each model for at least 10 epochs. We set the learning rate to 0.01 during the first 40 million batches, after which it is reduced to 0.001. The models are trained on CPUs across 50 machines using a distributed implementation of TensorFlow [1].

## 5.2 Offline Evaluation

Our models are evaluated offline on their ability to identify the true response to an email in the test data against a set of randomly selected competing responses. In this paper, we score a set of 100 responses that includes the correct response and 99 randomly selected incorrect competitors. We rank responses according to their scores, and report precision at 1 (*P@1*) as a metric of evaluation. We found that P@1 correlates with the quality of our models as measured in online experiments with users (see section 5.3).

| Batch Size | Scoring Model | P@1 |
|---|---|---|
| 25 | Joint | 49% |
| 25 | Dot-product | 48% |
| 50 | Dot-product | 52% |

Table 2. P@1 results on the test set for the joint and dot-product multi-loss scoring models. The training objective discriminates against more random negative examples for larger batch sizes.

Table 2 presents the results of the offline evaluation for joint and dot-product scoring models. The joint scoring model outperforms the dot-product model trained on the same batch size. This model learns complex cross-features between the input email and the response leading to a better scoring. However, the joint scoring model does not scale well to larger batches, since each possible pairing of input email and response requires a full forward pass through the network. The number of forward passes through the joint scoring model grows quadratically with the batch size. Recall the dot-product scoring model is a lot faster to train with multiple negatives than the joint scoring models since it requires a linear number of forward passes followed by a single $K$ by $K$ matrix multiply to score all possible pairings, where $K$ is the batch size. As a result, the multi-loss dot-product models can be trained on larger batches to produce more accurate models.

Note that the models in table 2 are trained with the multiple negatives training loss of section 4.4. It is also possible to train the models as a classifier with a sigmoid loss. We find that multiple negative training results in a consistent 20% reduction in error rate for P@1 relative to training as a classifier across all of our conversational datasets. For example, in a different version of the Smart Reply data it improved the P@1 of a dot-product model from 47% to 58%.

## 5.3 Online Evaluation

Though the offline ranking metric gives a useful signal during development, the ultimate proof of a response selection model is how it affects the quality of the suggestions that users see in the end-to-end Smart Reply system. Suggestion quality or usefulness is approximated here by the observed *conversion rate*, i.e. the percentage of times users click on one of the suggestions when they are shown.

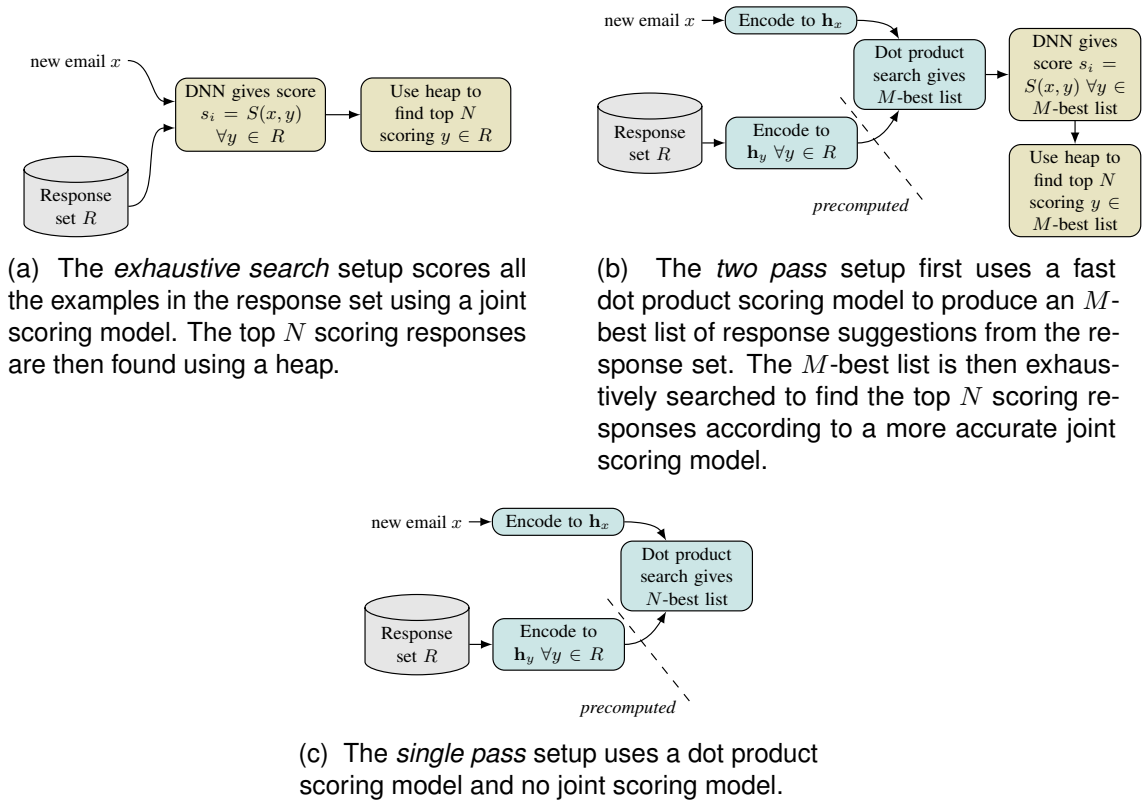

(a) The *exhaustive search* setup scores all the examples in the response set using a joint scoring model. The top $N$ scoring responses are then found using a heap.

(b) The *two pass* setup first uses a fast dot product scoring model to produce an $M$-best list of response suggestions from the response set. The $M$-best list is then exhaustively searched to find the top $N$ scoring responses according to a more accurate joint scoring model.

(c) The *single pass* setup uses a dot product scoring model and no joint scoring model.

Fig. 6. Online system architectures.

This section describes the evolution of our system and shows the effect of each iteration on latency and quality relative to the baseline Seq2Seq system. An outline of this series of online experiments is presented in table 3.

*5.3.1 Exhaustive Search.* Our initial system scored the input email against every response in the response set $R$ using the joint scoring model and the *Email body* feature alone (see figure 6a). Given that the joint scoring model requires a forward pass for each response in $R$, this approach is too computationally expensive for an online experiment, see row 1 of table 3.

*5.3.2 Two pass.* The computational expense of the initial exhaustive search system motivated the design of a two-stage approach where the first stage is fast and the second is more accurate, as shown in figure 6b.

The first stage consists of a dot-product scoring model utilizing the text of the email body alone. As a pre-processing step, all of the responses in the response set $R = \{y_1, \ldots, y_n\}$ are encoded to their vector representations to give a matrix $\mathbf{R} = [\mathbf{h}_{y_1}, \ldots, \mathbf{h}_{y_n}]$ (see figure 3b). At inference time, a new input email is encoded to its representation $\mathbf{h}_x$, and the vector of all scores is calculated as the dot product with the precomputed matrix: $\mathbf{R}\mathbf{h}_x$. A heap is then used to find the $M$ highest scoring responses. The second stage uses the joint scoring model to score the candidates from the first stage. Row 2 of table 3, shows the 50x speedup improvement from using this two pass system.

The system tended to suggest overly specific and often long responses because of the biased negative sampling procedure, see section 4.6. Therefore, we added an extra score to boost the scores of more likely responses using a language model. This change significantly improved the quality of the suggestions, see row 3 of table 3, moving the systems toward shorter and more generic responses that users were more likely to find appropriate and click.

| System | | Experiment | Conversion rate relative to Seq2Seq | Latency relative to Seq2Seq |
|---|---|---|---|---|
| Exhaustive search | (1) | Use a joint scoring model to score all responses in $R$. | – | 500% |
| Two pass | (2) | Two passes: dot-product then joint scoring. | 67% | 10% |
|  | (3) | Include response bias. | 88% | 10% |
|  | (4) | Improve sampling of dataset, and use multi-loss structure. | 104% | 10% |
| Single pass | (5) | Remove second pass. | 104% | 2% |
|  | (6) | Use hierarchical quantization for search. | 104% | 1% |

Table 3. Details of several successive iterations of the Smart Reply system, showing the *conversion rate* and latency relative to the baseline *Seq2Seq* system of Kannan et al. [11].

Improving our dataset sampling, and using the multi-loss structure brought the conversion rate of the system above that of the Seq2Seq system, see row 4 of table 3).

*5.3.3 Single pass.* To improve the latency, we removed the second pass step and relied solely on the responses found by the first pass dot-product step (see figure 6c). However, to maintain the quality, we had to improve the quality of the dot-product model.

Since the dot-product scoring model scales better with more negatives during training, we doubled the number of negatives for training the first pass system. We also applied the multi-loss architecture to the first pass dot-product model, using additional input features (see figure 4b). Together these changes made the dot-product model slightly more accurate than the joint model (see table 2). As a result, the system quality stayed the same while the speed increased 5 times, as shown in row 5 of table 3.

So far, we have been computing the dot-product between the new email representation and all the precomputed representations of the responses in the response set, and searching the entire list to find high scoring responses. Switching from this exhaustive search to the hierarchical quantization search described in section 4.7 doubles the speed of the system without compromising quality (see row 6 of table 3).

As a result, our final system produces better quality suggestions than the baseline Seq2Seq system with a small percentage of the computation and latency.

## 6 CONCLUSIONS

This paper presents a feed-forward approach for scoring the consistency between input messages and potential responses. A hierarchy of deep networks on top of simple n-gram representations is shown to outperform competitive sequence-to-sequence models in this context.

The deep networks use different components for reading inputs and precomputing the representation of possible responses. That architecture enables a highly efficient runtime search.

We evaluate the models with the Smart Reply application. Live experiments with production traffic enabled a series of improvements that resulted in a system of higher quality than the original sequence-to-sequence system and a small fraction of the computation and latency.

Without addressing the generation of novel responses, this paper suggests a minimal, efficient, and scalable implementation that enables many ranking-based applications.

## ACKNOWLEDGMENTS

Thanks to Fernando Pereira, Corinna Cortes, Anjuli Kannan, Dilek Hakkani-Tür and Larry Heck for their valuable input to this paper. We would also like to acknowledge the many engineers at Google whose work on the tools and infrastructure made these experiments possible. Thanks especially to the users of Smart Reply.